\documentclass[10pt,twocolumn,letterpaper]{article}

\usepackage{cvpr}              

\usepackage{graphicx}
\usepackage{amsmath}
\usepackage{amssymb}
\usepackage{booktabs}

\usepackage[pagebackref,breaklinks,colorlinks]{hyperref}

\usepackage[capitalize]{cleveref}
\crefname{section}{Sec.}{Secs.}
\Crefname{section}{Section}{Sections}
\Crefname{table}{Table}{Tables}
\crefname{table}{Tab.}{Tabs.}

\usepackage{multirow}

\newcommand{\hide}[1]{}

\def\cvprPaperID{9909} 
\def\confName{CVPR}
\def\confYear{2022}

\begin{document}



\title{Beyond Bounding Box: Multimodal Knowledge Learning for Object Detection}

\author{
Weixin Feng $^{1}$
\and
Xingyuan Bu $^{1}$\thanks{Corresponding author.}
\and
Chenchen Zhang $^{1}$
\and
Xubin Li $^{1}$
\and
$^1$Alibaba Group, China
\\
{\small
\texttt{ \{fengweixin.fwx,buxingyuan.bxy,zhangchenchen.zcc,lxb204722\}@alibaba-inc.com }
}
}

\maketitle

\begin{abstract}
Multimodal supervision has achieved promising results in many visual language understanding tasks, where the language plays an essential role as a hint or context for recognizing and locating instances. However, due to the defects of the human-annotated language corpus, multimodal supervision remains unexplored in fully supervised object detection scenarios. In this paper, we take advantage of language prompt to introduce effective and unbiased linguistic supervision into object detection, and propose a new mechanism called multimodal knowledge learning (\textbf{MKL}), which is required to learn knowledge from language supervision. Specifically, we design prompts and fill them with the bounding box annotations to generate descriptions containing extensive hints and context for instances recognition and localization. The knowledge from language is then distilled into the detection model via maximizing cross-modal mutual information in both image- and object-level. Moreover, the generated descriptions are manipulated to produce hard negatives to further boost the detector performance. Extensive experiments demonstrate that the proposed method yields a consistent performance gain by 1.6\% $\sim$ 2.1\% and achieves state-of-the-art on MS-COCO and OpenImages datasets.
\end{abstract}


\section{Introduction}
\label{introduction}


\begin{figure*}
\centering
\begin{subfigure}[htb]{0.95\linewidth}
\centering
\includegraphics[width=\linewidth]{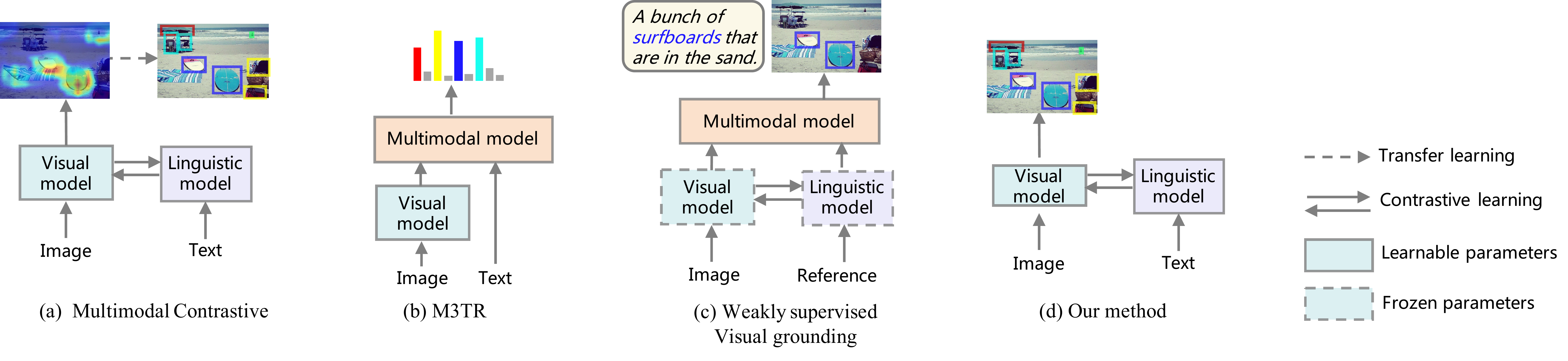}
\end{subfigure}
\caption{
Illustration of the proposed method. Previous methods provide evidence that multimodal knowledge can indirectly help recognize and locate instances. On the contrary, we propose to employ multimodal supervision directly into the detection model and design an end-to-end multimodal detector.
}
\vspace{-3mm}
\label{fig:intro_case}
\end{figure*}

Object detection is an essential and long-standing task that refers to recognizing and locating the instance within the desired categories. To tackle the difficulty, some works obtain high quality vision representations by designing fine structures such as feature pyramids~\cite{lin2017feature,ghiasi2019fpn}, cascade~\cite{cai2019cascade}, or transformer~\cite{carion2020end, wu2021visual}, other works instead remove the complexity from anchor by anchor-free detectors~\cite{zhou2019bottom,tian2019fcos,zhu2019feature}. Both of them aim to mine more information from the input image. 
However, the potential of supervision signals in other modalities is long neglected by the community. For instance, when humans intend to locate instances in the physical world, information in multiple modalities will be referred to. People will first \textit{listen to} the chirping then find the bird, and \textit{observe} its color and spots to judge the category. The \textit{smell} of cooking will help people to locate pots and foods. In a word, multimodal signals play an essential role in human reasoning. We argue that multimodal supervision can also facilitate the learning of detection models.

Recently, some works have indicated that multimodal supervision can support recognition and localization, as illustrated in Figure~\ref{fig:intro_case}. Self-supervised methods like Virtex~\cite{desai2021virtex} and Multimodal Contrastive~\cite{yuan2021multimodal} shown in Figure~\ref{fig:intro_case}(a) train a multimodal representation to embrace the global viewpoint from text, which can be transferred to detection task and surpass their visual counterpart.  M3TR \cite{zhao2021m3tr} in Figure~\ref{fig:intro_case}(b) indicates the capacity in multi-label recognition of linguistic knowledge through capturing the co-occurrence and context knowledge. Some works in weakly supervised grounding \cite{gupta2020contrastive, wang2021improving} learn visual-language alignment from frozen VL encoders. The image-level reference gives hint to the model to ground required objects, as illustrated in Figure~\ref{fig:intro_case}(c). In these works, the linguistic features that play a role of context or hint are trained together with the visual features to construct multimodal supervision and show promising results for enhancing recognition or localization. 


Despite the success of multimodal knowledge in these fields, using natural language directly in object detection task is non-trivial, which is mainly due to the defect of the image caption corpus. The human-annotated captions or descriptions are noisier and weaker than the bounding box annotation. For instance, humans usually tend to pay attention to the apparent objects while omitting the small and marginal ones, whereas these objects are the main difficulty lied in object detection. Secondly, a category can match extensive words in the descriptions owing to synonyms, leading to the difficulty of associating the language with the bounding boxes. Moreover, human annotation is expensive and hard to scale-up, which limits the model capabilities.  

To overcome the limitations of natural language and take full advantage of the pretrained language models, we propose a new mechanism called \textit{multimodal knowledge learning (MKL)} to directly incorporate the multimodal supervision into detection, as shown in Figure~\ref{fig:intro_case}(d). 
Inspired by the recent progress of language prompt~\cite{liu2021pre} in natural language processing, we first propose to use the language prompt that contains slots of categories, quantity, position, and relationships of objects in the image, to generate descriptions as multimodal supervision signal.
The prompt-based descriptions that integrate bounding box annotations into language form, are deemed to incorporate unbiased and more complete knowledge than image caption.
We further extract the context and co-occurrence information from the description via a large-scale pretrained language model. 
Then we maximize the mutual information of the representations in different modalities in a coarse-to-fine manner, where both image- and object-level supervision is applied to endow the detection model.
While the language description can be associated with the bounding box annotations, we can further apply prompt engineering to produce hard examples to train a more discriminative detector. This helps the detector focus on the false positive and false negative to further enhance the detection performance. 
The proposed method can be seamlessly integrated into any object detector. Moreover, the language model can be safely removed from the detector during the inference phase, incurring no more overhead than existing detectors at inference. Empirical results reveal that the proposed method achieves 1.6\% $\sim$ 2.1\% improvements on both COCO and OpenImages datasets.

In a nutshell, our main contributions are summarized as follows:
\begin{itemize}
\item We propose a novel method to utilize language description from prompt to provide multimodal supervision in object detection. And we demonstrate that the proposed method can be easily integrated into any detectors.
\item We further manipulate the generated description to produce hard negatives that force the detector more discriminative.
\item We provide detailed experiments of our method with different detection models on COCO and OpenImages datasets and achieve state-of-the-art.
\end{itemize}

\section{Related Work}

\paragraph{Generic Object Detection.}
In the era of deep learning, object detection is dominated by anchor-based methods, which can be generally divided into two-stage methods and one-stage methods~\cite{zhang2020bridging}.
Beginning with two-stage methods Faster R-CNN~\cite{ren2015faster} and its predecessors like Mask R-CNN~\cite{he2017mask} and Cascade R-CNN~\cite{cai2019cascade}, these methods use predefined anchors with different sizes and aspect ratios to generate the region of interests (RoIs), then refine the position and category of selected RoIs.
One-stage methods such as RetinaNet~\cite{lin2017focal} simplify the above pipeline by predicting the object category and box offsets directly from the anchor.
Recently, anchor-free methods receive more attention due to their flexibility.
Plenty of works are presented to discard the predefined anchor and turn to utilize the keypoints or center parts.
CornerNet~\cite{law2018cornernet} detects an object as two paired keypoints (top-left and bottom-right points).
GA-RPN~\cite{wang2019region} and FSAF~\cite{zhu2019feature} regard the center region of an object as positive. Meanwhile, FCOS~\cite{tian2019fcos} defines all the regions inside an object as positive.
However, these methods predict the category and bounding box individually from image modality without multimodal supervision.

\begin{figure*}[hbt]
\centering
\includegraphics[width=0.97\textwidth]{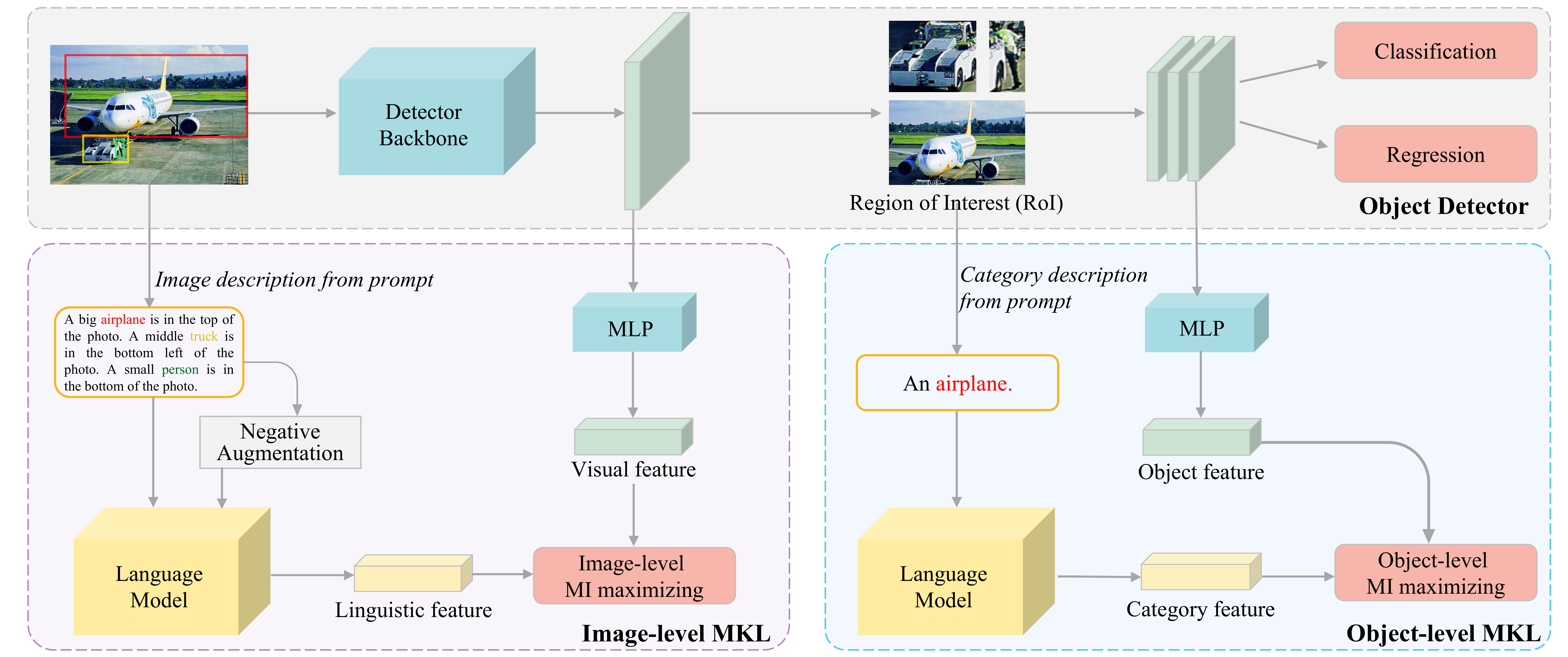}
\caption{The pipeline of our \textit{multimodal knowledge learning (MKL)} which can be integrated into any modern object detector. We generate language descriptions for both image- and object-level, then employ multimodal supervision to the detector through maximizing cross-modal mutual information.
Hard negative augmentation is used to generate negative features in image-level MKL to alleviate the false negatives and false positives of detection.}
\vspace{-3mm}
\label{fig:method}
\end{figure*}

\paragraph{Vision Language Understanding.}
Recent progress in VLU represents the potential of multimodal supervision in localization and recognition. 
Vision Language Understanding (VLU) aims to learn joint representation to accomplish various multimodal tasks, including visual question answering~\cite{antol2015vqa}, visual grounding~\cite{plummer2020revisiting, yang2019fast}, and image caption~\cite{stefanini2021show}. 
Typical works~\cite{li2019visualbert,li2020unicoder,lu2019vilbert,tan2019lxmert,li2020oscar,zhang2021vinvl} are armed with the capable transformer-based architectures~\cite{vaswani2017attention} to learn cross-modal interaction. 
Virtex~\cite{desai2021virtex} and Multimodal Contrastive~\cite{yuan2021multimodal} also learn multimodal knowledge for visual tasks like object detection and instance segmentation, which demonstrates its potential in localization through building global viewpoint.
Meanwhile, MDETR~\cite{kamath2021mdetr} proposes an end-to-end modulated detector to overcome the limitation of the pretrained detector in VLU. However, it mainly concentrates on the improvements on multimodal tasks. 
Another line of works aims to train network for a single multimodal task. The most relevant one to us is weakly supervised visual grounding. Wang \etal~\cite{wang2021improving} and Gupta \etal~\cite{gupta2020contrastive} maximize the mutual information between image and words. The words provide hints to the model for grounding the described object. However, both our technique and the task differ from them.
In addition, some Zero-shot learning works~\cite{li2021inference, rahman2020improved, gu2021zero} also introduce language supervision to recognize unseen categories.
VLU is also leveraged to multi-label~\cite{zhao2021m3tr, wang2020fast} recognition. These works capture the context and co-occurrence for recognition with the help of linguistic modality. On the contrary, our work further exploits the benefit of multimodal supervision for both localization and recognition. 


\paragraph{Prompt learning in NLP} Prompt learning paradigm~\cite{liu2021pre, brown2020language, petroni2019language} has rapidly aroused and attracted more and more interest in NLP field. Prompt learning is first used for knowledge probing~\cite{petroni2019language}, and then has been extended to numerous NLP tasks, such as question answering~\cite{khashabi2020unifiedqa}, natural language inference~\cite{schick2021exploiting}, commonsense reasoning~\cite{yang2020designing} and text generation~\cite{li2021prefix}. 
Prompt learning establishes a new direction for NLP. By training a language model with appropriate prompts of different NLP tasks, a single LM trained in an unsupervised fashion can be straightly adopted to solve numerous tasks without task-specific fine-tuning~\cite{liu2021pre, brown2020language, sun2021ernie}. Despite its popularity in NLP, prompt learning is barely explored in vision tasks. CLIP~\cite{radford2021learning} uses prompt to conduct zero-shot inference for the self-supervised learned model. In contrast, we show that prompt can also benefit the detection model in the training procedure.  

\section{Method}

\subsection{Overview}
The proposed \textit{multimodal knowledge learning (MKL)} can be adopted to different detection frameworks. Without loss of generality, hereafter we illustrate the overall structure of the proposed method using the two-stage detector framework as an example. The two-stage detector framework takes an image $I$ and its ground truth bounding boxes $B = \left\{b_{1},\ldots,b_{i},\ldots,b_{M}\right\}$ as input, where $b_{i} = \left(x_i,y_i,w_i,h_i,c_i \right)$ contains four coordinates $x_i,y_i,w_i,h_i$ and a class label $c_i$ of an object. In the first stage, the image $I$ is fed into a visual backbone such as ResNet to extract Regions of Interest (RoI) features $f^{r}$. In the second stage, RoI features are fed to classification and localization branches.

As illustrated in Figure~\ref{fig:method}, the proposed method consists of an \textit{Image-level MKL} in the first stage and an \textit{Object-level MKL} in the second stage. 
Both Image-level MKL and Object-level MKL benefit from multimodal supervision by generating language descriptions and maximizing the mutual information between visual and linguistic feature representations in contrastive learning. 
Specifically, Image-level MKL generates image-level descriptions from previously designed prompts to summarize the image and provide hints. Object-level MKL generates object-level descriptions that align the object RoI to incorporate fine-grained language information. Furthermore, augmented image-level prompts are used as \textit{hard negatives} to alleviate the common failures in detection.

\begin{figure*}[hbt]
\centering
\includegraphics[width=0.97\textwidth]{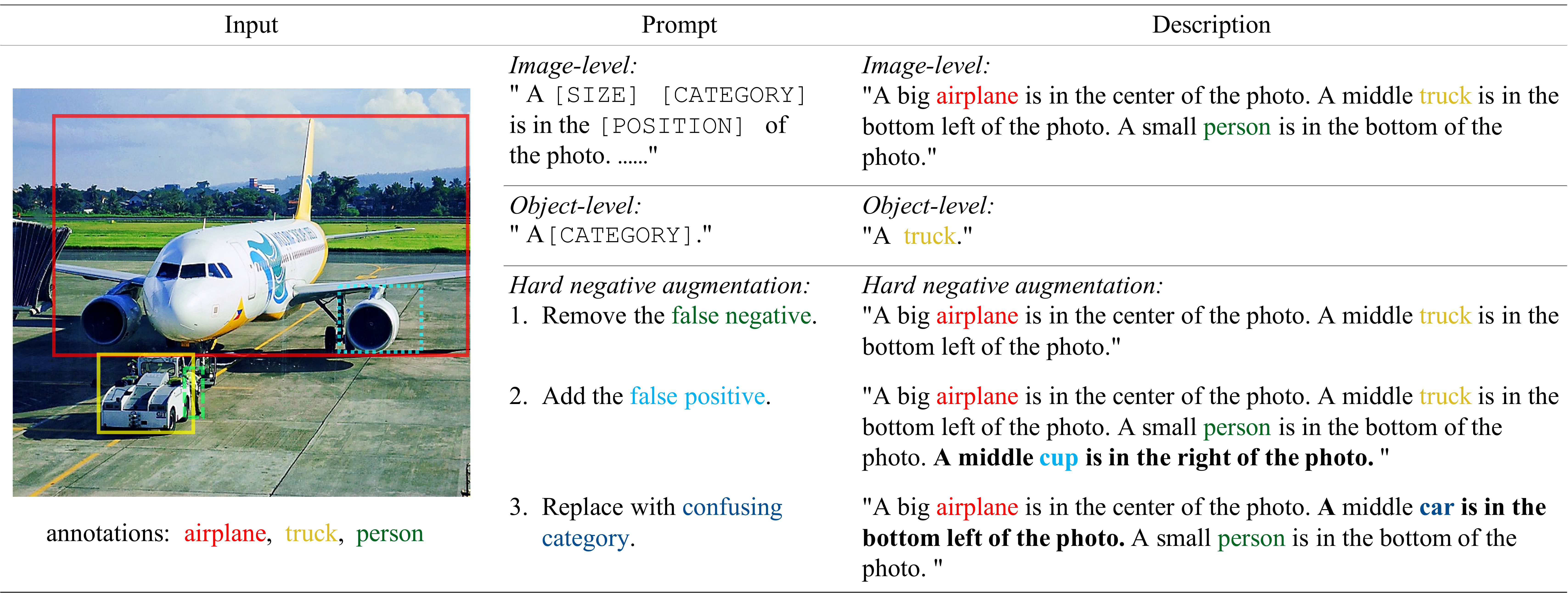}
\caption{Examples of different prompts in this work.}
\label{fig:prompt}
\end{figure*}

\subsection{Image-level MKL}

\paragraph{Prompt-based Description Generation.}
Although previous studies have achieved promising improvements on visual tasks by using linguistic supervision, the common corpus is weak and noisy so that not suitable in the fully supervised detection scenario. The human-labeled captions~\cite{chen2015microsoft} suffers from cognition bias that the unnoticeable and irregular objects tend to be ignored by the annotators. Another flaw is that the synonyms in natural language cause the difficulty of corresponding them to bounding boxes to provide fine-grained supervision. To provide effective and appropriate textual annotation, we use prompt~\cite{liu2021pre} to automatically generate image-level descriptions. The prompt-based description includes adequate information for recognition and localization, while is easy to scale-up. 

\hide{Although previous studies have achieved promising improvements on visual tasks by using language supervision, scarcity of textual annotations remains a severe bottleneck for fully supervised detection scenarios. Human-labeled captions~\cite{chen2015microsoft} are hard to scale-up, while webly-collected corpus~\cite{changpinyo2021conceptual} are noisy. For example, they tend to ignore unnoticeable and irregular objects due to the cognition bias. To provide a clean and complete textual annotation, we use prompt~\cite{liu2021pre} to automatically generate image-level descriptions, which is inexpensive.}

A prompt is a manually designed template that contains some slots, such as \texttt{[CATEGORY]}, \texttt{[POSITION]}, \texttt{[SIZE]}, and \texttt{[QUANTITY]}. Figure~\ref{fig:prompt} illustrates an example of the image-level prompt.
The prompt accepts the annotation $B$ as input and produces an image description $T$. More specifically, for an instance $b_{i} = \left(x_i,y_i,w_i,h_i,c_i \right) $, the \texttt{[CATEGORY]} is indicated by $c_i $,  the \texttt{[SIZE]} can be computed from $(w_i,h_i)$. For \texttt{[POSITION]}, we divide the image into 9 grids, each grid corresponding to a position.  The center coordinate $(\frac{x_i+w}{2},\frac{y_i+h}{2})$ is used to allocate the object into the 9 grids. Note that comparing with directly using $(\frac{x_i+w}{2},\frac{y_i+h}{2})$ to represent \texttt{[POSITION]}, discrete expression is closer to natural language and 
easy for model comprehension. The number of each category can be used to infer the \texttt{[QUANTITY]}. Finally, the sentences describing each object are concatenated to form $T$. We design diverse prompts to generate representative descriptions. Details of each prompt are documented in the supplementary material. \hide{The generated description will then be transformed to linguistic feature to introduce its semantic context into the object detector.}

\hide{Unlike those human-labeled captions~\cite{chen2015microsoft} and webly-collected corpus~\cite{changpinyo2021conceptual}, which are either hard to scale-up and tend to ignore unnoticeable objects due to the cognition bias, or noisy and irregular, our language prompt can give a clean and complete description of an image. Thanks to the prompt, our method can enhance the object detector on any detection datasets.}

\paragraph{Feature Extraction.}
To incorporate the language knowledge into detector, we first extract the \textit{linguistic feature} $f^{lg}$ from the image description $T $ and the \textit{visual feature} $f^{vg}$ from the visual backbone.
For the linguistic feature, we adopt a pretrained language model, such as BERT~\cite{devlin2018bert} or GPT-2~\cite{radford2019language}. The language model accepts a tokenized description as input, augments with position embedding, then produces a $d$-dimensional linguistic feature $f^{lg}$ from the \texttt{[EOS]} token.
For the visual feature, we fetch the feature map from the top of the backbone, a following global average pooling and a 2-layer MLP are used to construct $d$-dimensional visual feature $f^{vg}$.

\paragraph{Mutual Information Maximization.} We maximize the mutual information (MI) between linguistic feature and visual feature to incorporate the multimodal knowledge into the detection model. Mutual information is developed to measure the dependencies and correlation between two variables. Because MI is notoriously intractable to compute, we leverage an MI estimator InfoNCE~\cite{oord2018representation, chen2020simple} to estimate the lower bound of MI. Given a data batch with $N$ pairs of image-description, the objective for the $i$-th image is formulated as:
\begin{equation}
\mathcal{L}_{i}^{vg} = -log\frac{\exp(\langle f^{vg}_{i}, f^{lg}_{i}\rangle/\tau)}{\sum^{N}_{j=1}  \exp(\langle f^{vg}_{i}, f^{lg}_{j}\rangle/\tau)},
\label{eq:global_im}
\end{equation}
where $\langle \cdot \rangle$ represents the cosine similarity between two features, $\tau $ is a temperature hyper-parameter. By minimizing InfoNCE objective, the cross-modal mutual information can be enhanced. As a result, the knowledge existing in linguistic supervision can be distilled into the detector model to produce better representations. Meanwhile, we also maximize the MI in the textual side:
\begin{equation}
\mathcal{L}_{i}^{lg} = -log\frac{\exp(\langle f^{lg}_{i}, f^{vg}_{i}\rangle/\tau)}{\sum^{N}_{j=1}  \exp(\langle f^{lg}_{i}, f^{vg}_{j}\rangle/\tau)}.
\label{eq:global_txt}
\end{equation}
This symmetrized objective can help the pretrained language model adapt to the current prompt corpus.
Since the batch size per GPU in detection is relatively smaller than that of the classification task, we aggregate the visual and linguistic features over all devices during training.

\subsection{Object-level MKL}
The image-level description provides global linguistic supervision. However, the detector is not guaranteed to understand the relationship between each object and word. To fully improve the efficiency of multimodal supervision, we generate prompt-based object-level descriptions in object-level MKL.
As shown in Figure~\ref{fig:prompt}, each object-level prompt is designed to describe the category of an object. For each ground truth instance $b_{i} = \left(x_i,y_i,w_i,h_i,c_i \right) $, we obtain the \textit{object feature} $f^{vo}_{i}$ by using another 2-layer MLP which projects the RoI feature $f^r_i$ into a $d$-dimensional semantic space. We fill in the object-level prompt with the category $c_i$ to obtain the description $T_{c_i}$. Then, we use the same language model but armed with a different projection head on the last layer to produce a \textit{category feature} $f^{lo}_{c_i}$ with dimension $d$.


We further maximize the lower bound of MI between the object feature $f^{vo}$ and category feature $f^{lo}$. 
The objective is formulated as follows:
\begin{equation}
\mathcal{L}^{o} = -\sum\limits_{b_{i} \in B} log\frac{\exp(\langle f^{vo}_{i}, f^{lo}_{c_i}\rangle/\tau)}{\sum_{c_j =0}^{C}  \exp(\langle f^{vo}_{i}, f^{lo}_{c_j}\rangle/\tau)},
\label{eq:local_im}
\end{equation}
where $C $ is the category number in the dataset. The object feature $f^{vo}_{i}$ should discriminate the correct category feature $f^{lo}_{c_i}$ from the others, by which the detector can learn a finer alignment. We conduct this constraint for all the ground truth objects $b_i$ in $B $.
In addition, for the reason that the one-stage detectors perform classification and localization without RoIs, we only apply the object-level multimodal knowledge learning on two-stage detectors.

\subsection{Hard Negative Augmentation}
The image-level and object-level MKL associate the visual feature with linguistic supervision. In this section, we directly guide the detector to alleviate the failures drawing supports of prompt engineering~\cite{liu2021pre}. The prompt-based description has the advantage that the words can be associated with ground truth and the information can be flexibly inserted into or removed from it. Thus, we propose three augmentations on the prompt to generate hard negative examples from $T$, which can be applied to the image-level MKL. The first two augmentations focus on two common failures while the third augmentation aims to improve the robustness for confusing categories generally.

To achieve this, for each image $I $, we generate a set of hard negative examples as $\mathcal{H}_{t}$ with a fixed number $N_h $. As shown in Figure~\ref{fig:prompt}, the first two augmentations are to deal with the false negative and false positive predictions. 
We define these two cases by forwarding the detector and gathering the classification score $s \in \mathbb{R}^{C+1} $ for each prediction, where $C+1$ indicates the category number including background. Then we identify false negatives as those foregrounds with a smaller score than $\max(s)$. Likewise, we define false positives as those backgrounds with a smaller score than $\max(s)$.
For an object whose foregrounds are all fell into false negatives, we generate a hard negative via removing its description from $T$, and force the visual feature $f^{vg}$ to distinguish between the original and augmented description. Similarly, the second augmentation is to add the false positive prediction into $T$ when the failure occurs.
For the third augmentation, we randomly choose an object in the image and replace its category with a confusing category as a hard negative. 
In practice, we make use of the hierarchy label system in the WordNet~\cite{miller1995wordnet} and randomly choose a category with the same parent. 

We use the first two augmentations with high preference. 
Then we complement $\mathcal{H}_{t}$ with the third augmentation to obtain $N_h$ hard negative examples, which are then used jointly with $T $. The image-level visual constraint in Eq.~\ref{eq:global_im} can be reformulated as:
\begin{equation}
\mathcal{L}_{i}^{vg} = -log\frac{\exp(\langle f^{vg}_{i}, f^{lg}_{i}\rangle/\tau)}{\sum^{N+N_{h}}_{j=1}  \exp(\langle f^{vg}_{i} f^{lg}_{j}\rangle/\tau)}.
\label{eq:hard}
\end{equation}

\subsection{End-to-End Training}
The proposed \textit{multimodal knowledge learning} can be optimized jointly in an end-to-end manner with the object detector during training. Apart from the original detection loss $\mathcal{L}^{det}$, we introduce the image-level MI maximizing objective $\mathcal{L}^{vg}$ and $\mathcal{L}^{lg}$, and the object-level MI maximizing objective $\mathcal{L}^{o}$ to incorporate the language information into the detection model.
They are jointly optimized with the following loss:
\begin{equation}
\mathcal{L} = \mathcal{L}^{det} + 
\lambda^{vg} \mathcal{L}^{vg} +\lambda^{lg} \mathcal{L}^{lg} + \lambda^{o} \mathcal{L}^{o}.
\label{eq:overall}
\end{equation}
In the inference phase, the MKL module can be discarded. Therefore, our method does not increase any additional computational cost during inference, which illustrates its effectiveness and simpleness.

\section{Experiments}
\label{exp}

\subsection{Datasets}
We evaluate our method on the popular MS-COCO~\cite{lin2014microsoft} and OpenImages Challenge 2019~\cite{kuznetsova2020open}.
For the COCO dataset, we follow the standard train val split, which contains 118k training images and 5k validation images for 80 categories. 
If not specified, we report COCO style mAP which is the average AP over multiple overlap thresholds.
OpenImages dataset contains 500 hierarchical categories. 
Following the challenge 2019 protocol, we report AP$_{50}$ and use 1.7 million images as the train set, 40k images as the validation set.

\subsection{Implementation Details}
In all experiments, we adopt ResNet-50 as the visual backbone and use Faster R-CNN~\cite{ren2015faster} armed with FPN~\cite{lin2017feature} if not specified. For the language model, we use the same architecture as GPT-2~\cite{radford2019language}. We initialize the visual backbone from ImageNet pretrained model and initialize the language model from CLIP~\cite{radford2021learning}.
Considering the pretrained language model is trained on a more diverse corpus than our generated description, we only finetune the last projection layer. We then use 2-layer MLPs to transform features into $d=1024$ semantic space.
The number of hard negatives $N_h$ is set as 5, and the temperature $\tau $ is 0.07.
We use the mmdetection code base~\cite{mmdetection} and follow the default settings. The detectors are trained with 16 GPUs for COCO dataset and 64 GPUs for OpenImages. We adopt Tesla V100S-PCIE-32GB GPU with 4 images per device. Synchronized BatchNorm is used to make multi-GPU training more stable, and SGD is used to optimize the network for 13 epochs. The initial learning rate is set to 0.00125 per image and decreased by 0.1 after 8 and 11 epochs, respectively. The input images are resized so that the length of the shorter edge is 800 and the longer edge is limited to 1333.
In the joint optimization objective Eq.~\ref{eq:overall}, we use $\lambda^{vg}=0.5$, $\lambda^{lg}=0.5$, and $\lambda^{o}=0.1$ to balance the detection and MI maximizing tasks.

\begin{table}[t]
\centering
\begin{tabular}{cccc}
\toprule
Image-level & Object-level & Hard Negative & mAP \\
\midrule
 &  &  & 37.3 \\
\checkmark &  &  & 38.2 \\
\checkmark & \checkmark &  & 38.4 \\
\checkmark &  & \checkmark & 38.5 \\
\checkmark & \checkmark & \checkmark & 38.9 \\
\bottomrule
\end{tabular}
\caption{Results of each component in our proposed method.}
\label{table:abl}
\end{table}

\subsection{Ablation Study}
\label{sec:ablation}

\paragraph{Component-wise Analysis.} Firstly, we investigate the effect of the three main components in our design.
As shown in Table~\ref{table:abl}, the image-level MKL is shown effective which leads to a gain of 0.9\%. After being equipped with object-level supervision, our method boosts the performance by 1.1\%. And the hard negative augmentation contributes a further improvement of 0.5\% (38.4\% $\rightarrow$ 38.9\%). It is noted that simply adding hard negative augmentation only increases the AP by 0.3\% (38.2\% $\rightarrow$ 38.5\%) if without the help of object-level supervision. It is obvious that the detailed hints provided by object-level MKL are beneficial to hard negative augmentation.

\begin{table}[t]
\centering
\resizebox{0.90\linewidth}{!}{
\tabcolsep 0.06in{\scriptsize{}}%
\begin{tabular}{lcccccc}
\toprule
Method & AP & AP$_{50}$ & AP$_{75}$ & AP$_S$ & AP$_M$ & AP$_L$ \\
\midrule
Baseline & 37.3 & 58.0 & 40.2 & 21.5 & 40.8 & 48.1 \\
Word2vec & 37.7 & 58.8 & 40.8 & 21.1 & 41.7 & 48.4 \\
BERT    & 38.7 & 59.5 & 41.2 & 21.9 & 41.8 & 51.3 \\
GPT-2    & 38.6 & 59.5 & 40.8 & 22.4 & 42.0 & 51.3 \\
CLIP    & 38.9 & 59.9 & 41.6 & 22.3 & 42.2 & 51.2 \\
\bottomrule
\end{tabular}
}
\caption{Results of our proposed method with various language models.}
\label{table:lm}
\end{table}
\begin{table}[t]
\centering
\resizebox{1.0\linewidth}{!}{
\tabcolsep 0.01in{\scriptsize{}}%
\begin{tabular}{cccccc}
\toprule
\texttt{[CATEGORY]} & \texttt{[QUANTITY]} & \texttt{[POSITION]} & \texttt{[SIZE]} & \ Caption$^\dagger$ \ & \ mAP \ \\
\midrule
 &  &  &  &  & 37.3 \\
\checkmark & \checkmark &  &  &  & 38.4 \\
\checkmark & \checkmark & \checkmark &  &  & 38.7 \\
\checkmark & \checkmark & \checkmark & \checkmark &  & 38.9 \\
 &  &  &  & \checkmark & 37.8 \\
\bottomrule
\end{tabular}
}
\caption{Results of different image-level prompts. $^\dagger$ indicates only using COCO captions corpus.}
\label{table:prompt}
\end{table}

\begin{figure*}
\centering
\begin{subfigure}[b]{0.99\linewidth}
\centering
\includegraphics[width=\linewidth]{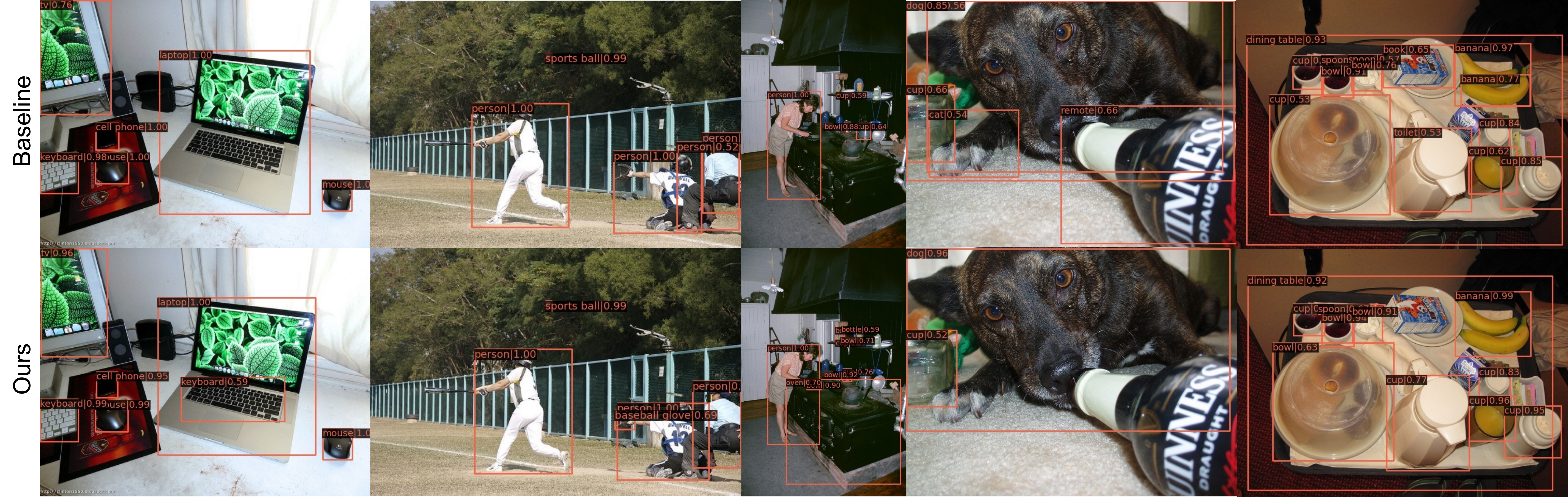}
\end{subfigure}
\caption{
Qualitative results on COCO val2017. The first two rows are predictions of Faster R-CNN baseline and Faster R-CNN armed with multimodal knowledge learning, respectively. multimodal knowledge learning alleviates the false negative in the first three columns and deposes the false positives in the last two columns.
}
\label{fig:exp_coco_case}
\end{figure*}

\begin{figure}
\centering
\includegraphics[width=0.95\linewidth]{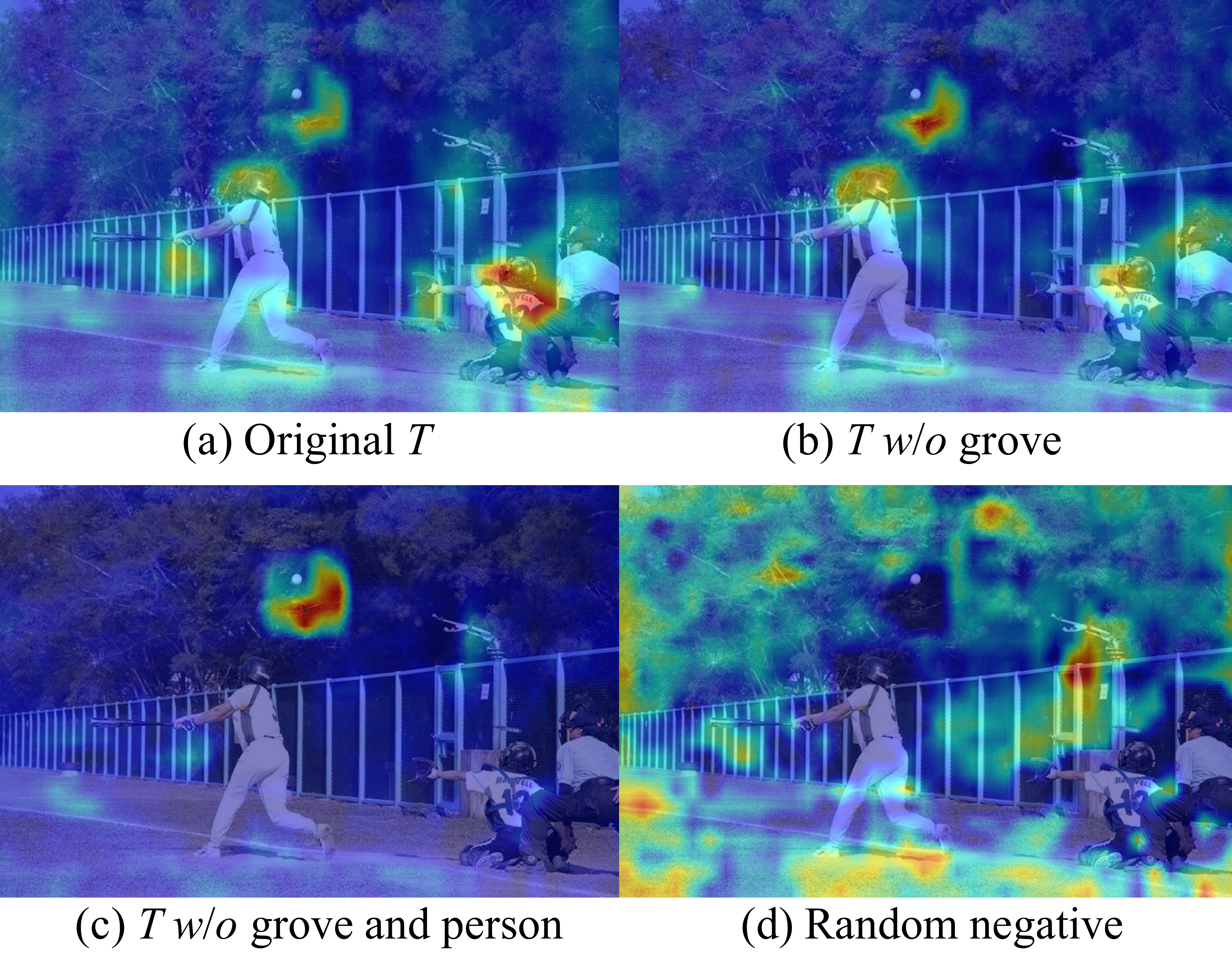}
\caption{
Visualization of the normalized response of different descriptions produced by Grad-CAM.
}
\label{fig:exp_cam}
\end{figure}

\paragraph{Comparison with Word2vec.} To evaluate the effectiveness of language prompt, we replace the prompt-based descriptions with word vectors. 
Specifically, we simply embedded the position and category information into word vectors via word2vec~\cite{rehurek_lrec} pretrained on google news. The simple word2vec can not fully leverage the rich hints and context knowledge in language, thus it can only provide limited improvement for 0.4\%, as shown in Table~\ref{table:lm}. 

\paragraph{Impact of Language Models.} Since the prompt is designed to fit the pretrained models, we discuss the impacts when using different language models in Table~\ref{table:lm}. We test two pretrained model BERT~\cite{devlin2018bert} and GPT-2~\cite{radford2019language} trained on HuggingFace~\cite{wolf-etal-2020-transformers}, which yield comparable gains of 1.4\% and 1.3\%, respectively. We also try a visual-language pretrained model from CLIP~\cite{radford2021learning}, where only the language part is used for a fair comparison. The VL pretrain is slightly better than the language pretrain, for it improves AP by about 1.6\%. In general, MKL can improve the detection model regardless of the choice of language model.

\hide{\paragraph{The Impact of Language Models.} Since a language model is involved in extracting the linguistic features, we study the impacts of different pretrained language models in Table~\ref{table:lm}.
We first test the simplest language model, \textit{i.e.}, word2vec~\cite{rehurek_lrec}, which is pretrained on google news. 
Word2vec only embeds words, thus we fed it category name to obtain language context. Results show that word2vec improves the AP by 0.7\%.
BERT~\cite{devlin2018bert} and GPT-2~\cite{radford2019language} pretrained models from HuggingFace~\cite{wolf-etal-2020-transformers} yields comparable gains of 1.4\% and 1.3\%, respectively.
CLIP~\cite{radford2021learning} is a multimodal model, we only use the language part for a fair comparison. With the help of the multimodal language feature, the AP improves by 1.6\%.}

\paragraph{Impact of Prompt Tuning.}
For the image-level MKL, we design different language prompts and randomly pick one of them for each image. The first prompt only holds the slots of \texttt{[CATEGORY]} and \texttt{[QUANTITY]}. Then we add the \texttt{[POSITION]} slot and design two more prompts. The \texttt{[SIZE]} slot is also joined in the other prompts. The details of different prompts are shown in the supplementary material. Table~\ref{table:prompt} discusses the results by using different prompts.
The category and quantity information can boost the performance by 1.1\%. The position and size information can also improve the object detector by assisting the localization.
We also attempt three orders of instances, \textit{i.e.}, random, sort by size, and default order, the results are float in 0.1\%. In practice, we adopt random order.
Moreover, we also instead of using COCO captions~\cite{chen2015microsoft} corpus, where only image-level supervision is employed. The COCO captions only gain the model for 0.5\% since it suffers from cognition error to provide strong supervision.


\paragraph{Visualization.}
Qualitative results are shown in Figure~\ref{fig:exp_coco_case}. We observe that multimodal knowledge learning can alleviate the false negatives and depose the false positives thanks to hard negative augmentation.
In Figure~\ref{fig:exp_cam}, we further illustrate the Grad-CAM~\cite{selvaraju2017grad} of the activations of different language descriptions. Figure~\ref{fig:exp_cam}(a $\sim$ c) show the results of the original description $T$ and augmented descriptions in Eq.~\ref{eq:hard}. We can see that all the objects are highlighted at the beginning, then the highlight regions vanish with the deletion of the corresponding object description. Clearly, the language description can perceive the location and category of instances in the image, which supports our hypothesis in Sec~\ref{introduction}. It also demonstrates that our language supervision can help the detector notice the given objects.
More visualization results are shown in the supplementary material.

\hide{In Figure~\ref{fig:exp_cam}, we further illustrate the Grad-CAM~\cite{selvaraju2017grad} for the activations to different descriptions. Figure~\ref{fig:exp_cam}(a $\sim$ c) shows the results of the original image-level description $T$, where all the objects are highlighted, indicating that our language description can capture high semantic context to endow detection. When removing some objects from the description, the corresponding highlight region vanish, as shown in Figure~\ref{fig:exp_cam}(b)(c), which shows the advantages of our method that it can help the network to notice a given object.}

\begin{table*}[t]
\centering
\resizebox{0.90\textwidth}{!}{
\begin{tabular}{lcccccccc}
\toprule
Method  & anchor type & MKL & AP & AP$_{50}$ & AP$_{75}$ & AP$_S$ & AP$_M$ & AP$_L$  \\
\midrule
\multicolumn{8}{l}{\textit{Two-stage detectors}} \\
\midrule
Faster R-CNN~\cite{ren2015faster}  &anchor-based&           & 37.3 & 58.0 & 40.2 & 21.5 & 40.8 & 48.1 \\
Faster R-CNN~\cite{ren2015faster}  &anchor-based& \checkmark & 38.9 & 59.9 & 41.6 & 22.3 & 42.2 & 51.2 \\
Mask R-CNN~\cite{he2017mask}   &anchor-based&           & 38.0 & 59.0 & 41.4 & 21.9 & 41.3 & 49.5 \\
Mask R-CNN~\cite{he2017mask}   &anchor-based& \checkmark & 39.1 & 60.5 & 42.0 & 22.3 & 42.2 & 51.9 \\
Cascade R-CNN~\cite{cai2019cascade}    &anchor-based&           & 40.4 & 58.7 & 43.8 & 22.7 & 43.6 & 53.4 \\
Cascade R-CNN~\cite{cai2019cascade}    &anchor-based& \checkmark & 41.4 & 60.2 & 44.7 & 23.0 & 44.4 & 54.9 \\
Deformable Faster R-CNN~\cite{zhu2019deformable}   &anchor-based&           & 41.2 & 62.6 & 45.0 & 24.8 & 44.5 & 54.1 \\
Deformable Faster R-CNN~\cite{zhu2019deformable}   &anchor-based& \checkmark & 42.3 & 63.9 & 46.0 & 25.1 & 45.9 & 56.8 \\
\midrule
\multicolumn{8}{l}{\textit{One-stage detectors }} \\
\midrule
RetinaNet~\cite{lin2017focal}  &anchor-based&           & 36.4 & 55.5 & 38.9 & 20.7 & 40.0 & 47.5 \\
RetinaNet~\cite{lin2017focal}  &anchor-based& \checkmark & 37.2 & 56.3 & 39.7 & 20.1 & 40.9 & 49.0 \\
GA-RetinaNet~\cite{wang2019region}  &anchor-free&           & 36.9 & 56.7 & 39.2 & 20.5 & 40.5 & 49.6 \\
GA-RetinaNet~\cite{wang2019region}  &anchor-free& \checkmark & 37.5 & 57.6 & 40.0 & 20.7 & 40.7 & 51.5 \\
FSAF~\cite{zhu2019feature} &anchor-free&           & 37.0 & 56.4 & 39.5 & 20.6 & 40.6 & 48.3 \\
FSAF~\cite{zhu2019feature} &anchor-free& \checkmark & 37.8 & 57.2 & 40.2 & 21.1 & 40.8 & 49.6 \\
\bottomrule
\end{tabular}
}
\caption{Results of our proposed method with various detectors. MKL means multimodal knowledge learning.}
\label{table:various_detectors}
\end{table*}

\begin{table}[t]
\centering
\resizebox{0.45\textwidth}{!}{
\tabcolsep 0.07in{\scriptsize{}}
\begin{tabular}{lcccc}
\toprule
\multirow{2}{*}{Backbone}  & \multirow{2}{*}{MKL} & \multirow{2}{*}{AP} & \multicolumn{2}{c}{runtime} \\
\cmidrule{4-5}
                            &                       &                   & train & test \\
\midrule
ResNet-50  &           & 54.5  & 130.3ms & 70.7ms  \\
ResNet-50  & \checkmark & 56.6 & 152.8ms & 73.1ms  \\
\midrule
ResNet-101  &           & 57.0 & 176.6ms & 76.4ms  \\
ResNet-101  & \checkmark & 58.4 & 196.1ms & 76.5ms \\
\midrule
ResNet-152  &           & 58.4 & 233.9ms & 89.2ms \\
ResNet-152  & \checkmark & 60.4 & 255.3ms &  88.8ms \\
\midrule
ResNeXt-32x4d-50  &   & 55.7 & 153.3ms & 78.2ms \\
ResNeXt-32x4d-50  & \checkmark & 57.6 & 177.4ms & 76.0ms \\
\midrule
ResNeXt-32x8d-101  &   & 59.7 & 322.4ms & 118.7ms \\
ResNeXt-32x8d-101  & \checkmark & 61.0 & 343.0ms & 119.3ms \\
\bottomrule
\end{tabular}
}
\caption{Results on OpenImages dataset with various backbones. MKL means multimodal knowledge learning. The runtime is the average value of 1000 iterations on a single Tesla V100S-PCIE-32GB.}
\label{table:various_datasets}
\end{table}

\subsection{Applicable to State-of-the-art Detectors}
In this section, we conduct a quantitative comparison of our method with other state-of-the-art detectors.
Due to the inherent properties of the proposed method, it is applicable to other detectors including one-stage methods and two-stage methods.
For two-stage methods, all settings are as discussed above.
For one-stage methods, we use half the learning rate of two-stage methods and omit the object-level supervision due to their dense prediction.
As shown in Table~\ref{table:various_detectors}, the proposed method is widely applicable to various object detection methods.
In particular, our multimodal knowledge learning boosts the two-stage detectors by 1.0\% $\sim$ 1.6\% even in the strong baseline of Deformable Faster R-CNN~\cite{zhu2019deformable}.
Considering the object-level supervision is omitted in the one-stage detection family, the gain is lower than that of two-stage methods. Nevertheless, our method can improve the anchor-based RetinaNet~\cite{lin2017focal} and anchor-free GA-RetinaNet~\cite{wang2019region} and FSAF~\cite{zhu2019feature} consistently. The results on different state-of-the-art provide strong evidence of the effectiveness of multimodal supervision.

\hide{We find that the improvement on Faster R-CNN~\cite{ren2015faster} is more significant than that on Mask R-CNN~\cite{he2017mask}, Cascade R-CNN~\cite{cai2019cascade}, and Deformable Faster R-CNN~\cite{zhu2019deformable}. We argue the reason is that the latter methods leverage other cues, such as instance segmentation or deformable convolution, to enhance the context.}

\subsection{Generalization on Large-scale Detection}
Since multimodal knowledge learning has demonstrated outstanding performance and generalization on various state-of-the-art methods on the COCO dataset, we further corroborate the proposed method on the large-scale OpenImages dataset.
Based on Faster R-CNN, various backbones are adopted to fully validate the generalization of multimodal knowledge learning.
Table~\ref{table:various_datasets} summarizes the detailed results on OpenImages. Multimodal knowledge learning steadily improves the performance by 2.0\% $\sim$ 2.1\% across various backbones.
Note that the proposed method yields a performance gain of 2.1\% with ResNet-50 on OpenImages, which is more significant than that on the COCO dataset. 
This comparison indicates that with a more complex hierarchy category system on OpenImages, the detection model will capture richer context and co-occurrence knowledge from maximizing the cross-modal mutual information. 
In addition, adding MKL only introduces a small computational overhead during training and basically does not bring any extra cost during inference. The running time shows the feasibility and economy of the proposed method.

\section{Conclusion}
In this paper, we propose a novel mechanism named \textit{multimodal knowledge learning}, which incorporates multimodal supervision to guide the detection model. It generates language descriptions from existing bounding box annotation via prompt as linguistic supervision, then its knowledge is leveraged to the detection model via maximizing cross-modal mutual information on both image-level and object-level.
The self-generated prompt makes the proposed method pluggable for various datasets even without human-labeled captions, and the experiments show that we achieve consistent improvements on COCO and OpenImages datasets. Our method can also be applied to various detectors and improve their performance. 
We hope this work can open up new opportunities to explore multimodal detection frameworks rather than solely relying on visual information.

{\small
\bibliographystyle{ieee_fullname}
\bibliography{egbib,main}
}

\end{document}